%% file: main.tex
\documentclass[sigconf]{acmart}
\usepackage{amsmath}
\usepackage{amsfonts}
\usepackage{algorithmic}
\usepackage{graphicx}
\usepackage{textcomp}
\usepackage{xcolor}
\usepackage{dsfont}
\usepackage{multirow}
\usepackage{subcaption}
\usepackage{xurl}

\AtBeginDocument{%
  \providecommand\BibTeX{{%
    \normalfont B\kern-0.5em{\scshape i\kern-0.25em b}\kern-0.8em\TeX}}}

\acmYear{2025}\copyrightyear{2025}
\setcopyright{cc}
\setcctype[4.0]{by-nd}
\acmConference[BuildSys '25]{The 12th ACM International Conference on Systems for Energy-Efficient Buildings, Cities, and Transportation}{November 19--21, 2025}{Golden, CO, USA}
\acmBooktitle{The 12th ACM International Conference on Systems for Energy-Efficient Buildings, Cities, and Transportation (BuildSys '25), November 19--21, 2025, Golden, CO, USA}
\acmDOI{10.1145/3736425.3770098}
\acmISBN{979-8-4007-1945-5/25/11}
\begin{document}


\title{MUDAS: Mote-scale Unsupervised Domain Adaptation in Multi-label Sound Classification}


\author{Jihoon Yun}
\affiliation{
  \institution{The Ohio State University}
  \city{Columbus}
  \state{OH}
  \country{USA}}
\email{yun.131@osu.edu}

\author{Chengzhang Li}
\affiliation{
  \institution{The Ohio State University}
  \city{Columbus}
  \state{OH}
  \country{USA}}
\email{li.13488@osu.edu}

\author{Dhrubojoyti Roy}
\affiliation{%
  \institution{Microsoft}
  \city{Redmond}
  \state{WA}
  \country{USA}}
\email{dhrroy@microsoft.com}

\author{Anish Arora}
\affiliation{%
  \institution{The Ohio State University}
  \city{Columbus}
  \state{OH}
  \country{USA}}
\email{arora.9@osu.edu}


\begin{abstract}
Unsupervised Domain Adaptation (UDA) is essential for adapting machine learning models to new, unlabeled environments where data distribution shifts can degrade performance.
Existing UDA algorithms are designed for single-label tasks and rely on significant computational resources, limiting their use in multi-label scenarios and in resource-constrained IoT devices. 
Overcoming these limitations is particularly challenging in contexts such as urban sound classification, where overlapping sounds and varying acoustics require robust, adaptive multi-label capabilities on low-power, on-device systems. 
To address these limitations, we introduce Mote-scale Unsupervised Domain Adaptation for Sounds (MUDAS), a UDA framework developed for multi-label sound classification in resource-constrained IoT settings.
MUDAS efficiently adapts models by selectively retraining the classifier in situ using high-confidence data, minimizing computational and memory requirements to suit on-device deployment. 
Additionally, MUDAS incorporates class-specific adaptive thresholds to generate reliable pseudo-labels and applies diversity regularization to improve multi-label classification accuracy. 
In evaluations on the SONYC Urban Sound Tagging (SONYC-UST) dataset recorded at various New York City locations, MUDAS demonstrates notable improvements in classification accuracy over existing UDA algorithms, achieving good performance in a resource-constrained IoT setting.


\end{abstract}




\maketitle

\section{INTRODUCTION}
\input{sections/introduction}

\section{RELATED WORKS}
\input{sections/related_works}

\section{SYSTEM MODEL \& PROBLEM STATEMENT}
\input{sections/system_model}

\section{UNSUPERVISED DOMAIN ADAPTATION IN MULTI-LABEL CLASSIFICATION}
\input{sections/UDA}

\section{ON-DEVICE RESOURCE-EFFICIENT RELEARNING}
\input{sections/relearning}

\section {EVALUATION}
\input{sections/evaluation}

\section{CONCLUSIONS}
\input{sections/conclusion}

\vspace{-4pt}
\bibliographystyle{ACM-Reference-Format}
\bibliography{main}


\end{document}

%% file: sections/introduction.tex
UDA is a powerful machine learning technique that enables models to transfer knowledge from a labeled source domain to an unlabeled target domain \cite{Domain_Adversarial_1, Domain_Adversarial_2, MMD_1, Ensembling}. 
In real-world applications, data distribution often varies across environments---a phenomenon known as domain shift. 
UDA addresses this issue by adapting models to new environments without requiring additional labeled data from the target domain. 
It has been widely used to improve model generalization in a variety of settings, particularly in areas where labeled data is scarce or difficult to obtain.

Existing UDA algorithms, such as AdaMatch \cite{AdaMatch}, are primarily designed to handle single-label classification tasks. 
In these tasks, each data point is assigned exactly one label, making it relatively straightforward to train models.
However, real-world scenarios, such as urban sound classification, often involve multi-label classification. 
In these tasks, a single instance can belong to multiple classes simultaneously, such as when an audio clip contains overlapping sounds like human voices, emergency sirens, and industrial machinery. 

Additionally, existing UDA algorithms often require substantial computational resources, which makes them unsuitable for deployment on small IoT devices that are commonly used in smart cities.
Thus, even though UDA is in principle well suited for adaptation of models that run on devices and even with the relearning done in situ in an online fashion, the constraints of these devices---such as limited memory, processing power, and battery life---pose further challenges for the practical application of UDA. 
These devices, often referred to as ``mote-scale'' devices, are deployed in large numbers and are expected to run continually with minimal maintenance. 
For IoT applications, such as urban sound classification, which is the representative application we focus on in this paper, a model and its relearning in the device must be efficient enough to run together on these devices while still maintaining high performance despite the domain shifts encountered by the device by virtue of its environment changing with respect to location or time. 
Traditional UDA approaches do not adequately address the unique needs of multi-label tasks in these resource-constrained settings.

\begin{figure}[t]
\centering
\includegraphics[width=0.25\textwidth, trim=9 0 0 0, angle=-90]{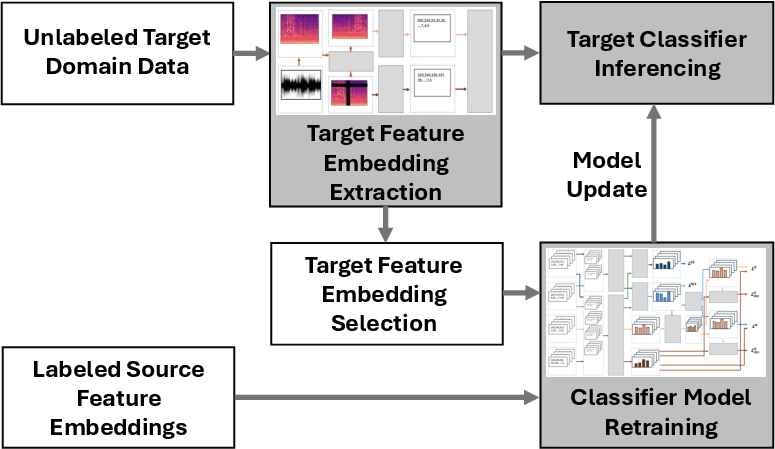}
\vspace{-4pt}
\caption{The MUDAS Framework}
\label{fig:structure}
\vspace{-14pt}
\end{figure}

To overcome these limitations, we introduce MUDAS, a novel framework specifically designed to address multi-label classification in IoT environments under unsupervised domain adaptation, and use MUDAS to solve an urban sound classification problem.
The framework of MUDAS is shown in Figure \ref{fig:structure}.
In MUDAS, audio data from deployed IoT devices in the target domain is processed to extract feature embeddings and generate inferences. 
High-confidence embeddings are stored for future classifier retraining, balancing adaptation with resource constraints.

MUDAS incorporates several advanced techniques to improve multi-label classification in unsupervised settings. 
One of the key components is the use of class-specific adaptive thresholds to generate high-confidence pseudo-labels, which helps to mitigate the noise inherent in unsupervised label generation. 
Traditional approaches often use a single threshold for pseudo-labeling, but this is insufficient for multi-label classification, where each label can have a different level of confidence. 
By introducing both positive and negative thresholds tailored to each class, MUDAS ensures that the generated pseudo-labels are more accurate and reliable, which is critical for improving performance in real-world applications. 
Furthermore, MUDAS applies diversity regularization to reduce the risk of overfitting,  improve the robustness of the model, and incorporate features for each class ---as opposed to frequently occurring ones--- when adapting to new, unlabeled data.

MUDAS also addresses the challenge of resource-efficient on-device adaptation by leveraging high-confidence data for periodic retraining. 
Rather than continuously retraining the model on every new data point, MUDAS stores only high-confidence embeddings and retrains the classifier periodically when sufficient data has been collected. 
This approach conserves both memory and computational resources, ensuring that IoT devices can maintain high performance while minimizing energy consumption. 

To evaluate MUDAS, we use the SONYC-UST dataset \cite{SONYC_UST_V1},\cite{SONYC_UST_V2}, which contains sound recordings labeled across multiple classes and recorded at various New York City locations. We simulate real-world domain shifts by splitting the dataset by location, treating one location as the source domain and another as the target. MUDAS is benchmarked against two baselines: a lower-bound model trained only on source data and tested on target data, and an upper-bound model trained and tested on target data. Results demonstrate that MUDAS outperforms these baselines, achieving good classification performance in resource-constrained, mote-scale environments.

We summarize the main contributions of this papers as follows:
\vspace*{.25mm}
\begin{itemize}
    \item We introduce MUDAS, a novel framework tailored to address the unique challenges of multi-label classification in unsupervised domain adaptation.  
    Compared with the existing AdaMatch framework, which is optimized for multi-class classification tasks, MUDAS incorporates three key innovations---class-specific adaptive thresholds, inclusion of positive and negative thresholds, and diversity regularization---to address the challenges of multi-label classification effectively.
    \item We explore the deployment of MUDAS on resource-constrain-ed edge devices by optimizing its computational and storage efficiency. 
    For computational efficiency, MUDAS performs feature embedding inference directly on edge devices and retrains only the downstream classifier, significantly reducing processing demands. 
    To address storage limitations, we propose an algorithm that selectively store high-confidence unlabeled target data for retraining while discarding less relevant data, ensuring efficient use of limited device storage.
    \item  We evaluate MUDAS extensively on the SONYC-UST dataset, which includes sound recordings from diverse New York City locations. The results demonstrate that MUDAS achieves notable (up to 6.7\%) improvements in classification accuracy relative to the source model.
\end{itemize}

%% file: sections/related_works.tex
\subsection{Multi-label Sound Classification}
In sound processing, multi-label classification frequently arises due to overlapping audio sources in real-world environments. Pre-trained neural network embeddings like VGGish \cite{VGGish} and Open$L^3$ \cite{L3_NET} capture high-level audio features and model complex acoustic patterns, improving the classification of concurrent sound events. These embeddings reduce dimensionality, enhance computational efficiency, and preserve essential sound characteristics, which boosts generalization and resilience to background noise across diverse environments. However, their large model sizes limit deployment on resource-constrained devices. To address this, methods such as \cite{VGGISH_IOT}, \cite{BNN}, \cite{Edge_L3}, \cite{SEA}, and \cite{MKII} optimize these models for edge computing. Despite advances in embedding-based classification, a significant challenge remains: enabling on-device classifier adaptation to domain shifts, particularly in resource-limited settings.

\vspace{-8pt} 

\subsection{Unsupervised Domain Adaptation}
UDA enables models to transfer knowledge from a labeled source domain to an unlabeled target domain. A popular approach in UDA is Domain-Adversarial Training, which uses a domain discriminator to align features between source and target domains, facilitating domain-invariant feature learning and enhancing model robustness \cite{Domain_Adversarial_1}. Another technique, Maximum Mean Discrepancy (MMD), reduces the domain gap by aligning the statistical means of feature distributions between source and target domains \cite{MMD_1, Domain_Adversarial_2}. Self-Ensembling, on the other hand, employs a “teacher” model to generate pseudo-labels for target data, guiding a “student” model in adapting to the target domain without requiring labeled target data \cite{Ensembling}.

AdaMatch \cite{AdaMatch} is a state-of-the-art method for UDA and semi-supervised learning. It achieves distribution alignment between source and target domains using batch normalization statistics and employs confidence-thresholded pseudo-labeling to enhance adaptation performance. While AdaMat-ch has proven highly effective for multi-class image classification, applying its techniques to other domains presents unique challenges.

Our study extends the principles of AdaMatch to the domain of multi-label urban sound classification. Unlike multi-class tasks, multi-label classification often involves instances with overlapping labels, which makes direct adaptation approaches less effective without adjustments tailored to the unique characteristics of sound data. This work adapts these techniques to address the complexities of urban sound datasets, including label overlap and environmental noise.

In some scenarios, multi-class classification can outperform multi-label methods in urban sound classification. This occurs when dataset characteristics—such as an uneven distribution of labels in specific locations or a sparse occurrence of multiple labels—diminish the advantages of a multi-label approach. Unlike curated datasets, real-world urban sound data often reflects natural imbalances and uncontrolled label generation. In such cases, dominant labels or location-specific trends can bias the dataset, making multi-class classification more effective by focusing on distinct patterns associated with the dominant labels. 

\vspace{-8pt} 

\subsection{On-Device Learning}
To prevent domain shifts, models need to be trained on data specific to the deployment location. However, collecting such data offline from all deployment locations is impractical, making on-device learning essential.

Advancements in microcontroller technology, such as the ARM Cortex-M Series \cite{M7}, STM32H7 \cite{STM32H7}, and NXP i.MX RT series \cite{iMX_RT}, have made it possible to deploy machine learning models in resource-constrained environments. Supporting these developments, libraries like CMSIS-NN \cite{CMSIS_NN}, LiteRT \cite{LiteRT}, and MCUNet \cite{MCUNet} optimize model inference for embedded systems. In addition, tools like TTE \cite{256KB}, POET \cite{POET}, and MiniLearn \cite{MiniLearn} enhance on-device training capabilities, even for highly resource-constrained mote-scale devices. 
While these technological advancements are promising, the application of UDA for multi-label sound classification in highly resource-constrained, on-device environments has seen little exploration, leaving a significant gap in current research.

%% file: sections/system_model.tex
Consider a sound classification scenario in a wide area urban environment where numerous acoustic devices capture audio clips from their local surroundings. 
Each device must classify the recorded audio into multiple labels---such as human voice, engine noise, dog barking, etc. 
Notably, each audio clip can contain zero, one, or multiple classification labels; for example, both human voice and engine noise might be present in a single clip. Therefore, this is a \emph{multi-label} classification problem.

Each device employs a locally-stored neural network for classification, trained and fine-tuned using labeled data collected by the devices themselves. 
The labels are provided by human annotators. This approach \cite {SONYC_UST_V1, alias2019review} has been used successfully to deploy cyberphysical systems where environmental monitoring is used to guide noise enforcement or  remediation \cite{bello2019sonyc}, analyze human activities, and recognize machine failure conditions.

In this paper, we study an \emph{unsupervised domain adaptation} problem where the goal is to classify sounds in a new environment, specifically when devices are deployed in a different geographic location.
For the new environment, labeled data is unavailable due to the absence of human annotators. 
Thus, unsupervised domain adaptation is needed to adapt the pre-trained neural network to the new environment.

We define the original environment as the ``source domain'' and the new environment as the ``target domain''.
Our study focuses on leveraging labeled data from the source domain and unlabeled data from the target domain to learn a model that performs well on the target domain’s data.

Formally, the objective function in this paper is defined as:
\vspace{-0.5em}
\begin{equation}
J=\max_{\theta} \left| \text{AUPRC}^{t}( f_{\theta}(x^s_{ [1..k]}, y^s_{[1..k]}, x^t_{[1..k]})) \right|
\end{equation}

AUPRC (Area Under the Precision-Recall Curve) is a metric used to evaluate classification performance, particularly in scenarios with class-imbalanced datasets. In this context, the AUPRC score is calculated for the test set of the target domain, denoted as $\text{AUPRC}^{t}$.
The model $f_{\theta}(x^{s}_{[1..k]}, y^{s}_{[1..k]}, x^t_{[1..k]})$ is trained using source domain data points $x^{s}_{[1..k]}$ with corresponding labels $y^{s}_{[1..k]}$, and unlabeled target domain data points $x^t_{[1..k]}$, across all classes 1 to $k$.
It is important to note that the source domain dataset $x^{s}_{[1..k]}$ and the target domain dataset $x^t_{[1..k]}$ may differ in size, with $n$ and $m$ data points, respectively. The goal of this paper is to maximize $J$.

It is important to note that the domain adaptation is performed \emph{on-device}, meaning the model is relearned directly on edge devices using selected unlabeled target data along with labeled source data.
Due to the limited resources available on these edge devices, our algorithm must be designed to meet the constraints on memory and computational capacity, ensuring efficient performance during the relearning process.

%% file: sections/UDA.tex
\begin{figure*}[t]
\centering
\includegraphics[width=0.365\textwidth, trim = 9 0 0 0, angle=-90]{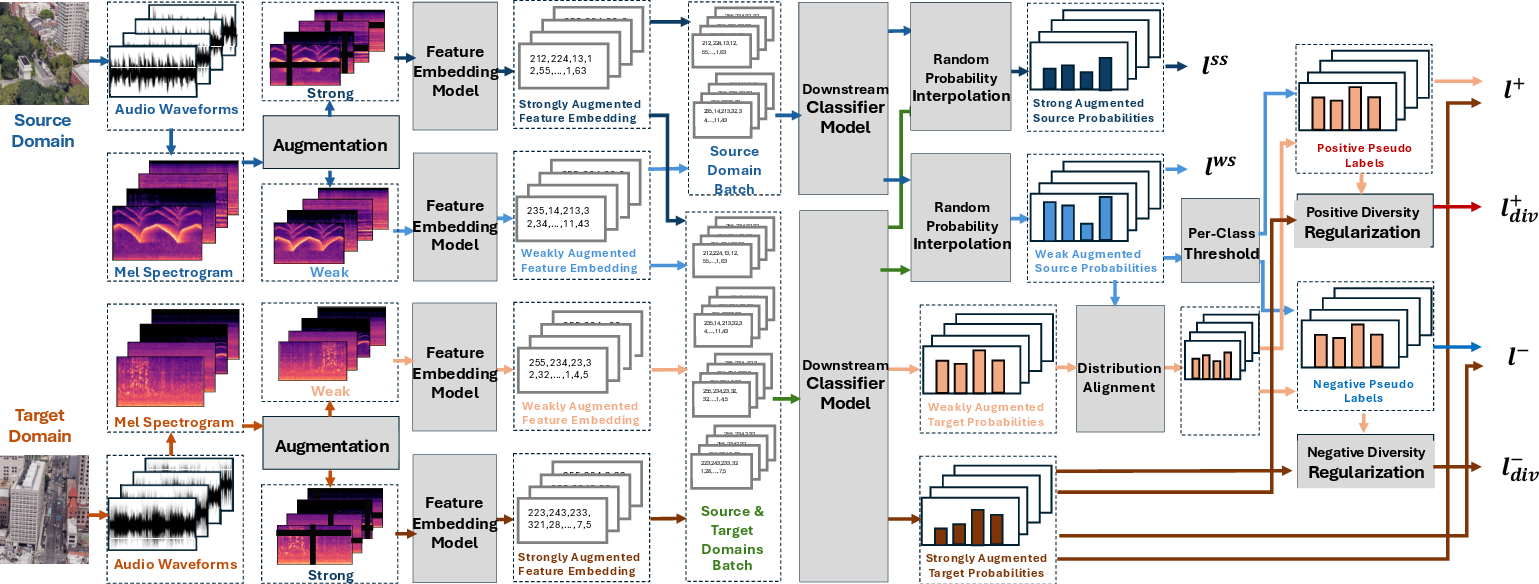}
\vspace{-4pt}
\caption{Flow of computation in MUDAS training}
\vspace{-10pt}
\label{fig:flow}
\end{figure*}

In this section, we describe the elements of MUDAS; cf.~its computational flow illustrated in Figure \ref{fig:flow}. 
Our approach use a labeled data set from the source domain and an unlabeled data set from the target domain, both consisting of audio waveforms. 
We utilize a pre-trained neural network to transfer the audio waveforms into \emph{feature embeddings}, which are high-level representations extracted from the datasets. For feature embedding generation, we use the SONYC-L\textsuperscript{3} \cite{MKII} model that is optimized for embedding extraction and inference on mote-scale devices. We train our classifier NN based on the SONYC-L\textsuperscript{3} feature embeddings (labeled from the source domain and unlabeled from the target domain). 

For the training of the classifier, we create two separate batches: one containing feature embeddings solely from the source domain and another incorporating feature embeddings from both source and target domains. 
This allows the model to better adapt to the target domain while preserving the integrity of the source domain’s features.

To further improve the model’s representational capacity, we adopt random probability interpolation from the AdaMatch framework. 
This technique merges the batch statistics of the source and target domains, which fosters the development of a more diverse feature space, contributing to the model’s enhanced ability to generalize across domains. 
We also adopt distribution alignment from the AdaMatch framework to refine the predicted class distributions, aligning them more closely with the expected true distribution. This step promotes consistent predictions across both source and target domains, thereby mitigating the effects of domain shift.

We introduce distinct, class-specific positive and negative thresholds to generate pseudo-labels from weakly augmented target probabilities. Unlike AdaMatch, which is designed for multi-class classification, our approach addresses the unique challenges of multi-label classification, where selecting a single label based on relative confidence is insufficient. These thresholds are calibrated to align closely with the distribution of weakly augmented source probabilities, ensuring a meaningful basis for pseudo-labeling. Given the tailored pseudo-labels, we then apply positive and negative diversity regularization within the loss function. This mechanism strengthens the model’s robustness, driving it to produce diverse yet consistent predictions across domains, thereby enhancing its resilience and adaptability to varying target distributions.
 
\noindent\textbf{Notation} Let $x$, $y$, and $z$ represent the embeddings, labels, and predicted probabilities, respectively. 
In the source domain, we have $n$ embeddings $x^{s}$ with their corresponding labels $y^{s}$. In the target domain, we have $m$ embeddings $x^{t}$, but the labels $y^{t}$ are not available. During each training step, we use $b^{s}$ and $b^{t}$ batches from the $n$ source and $m$ target embeddings, respectively. Each embedding is augmented both strongly and weakly. We denote the strongly and weakly augmented source embeddings as $x^{ss}$ and $x^{ws}$, and the strongly and weakly augmented target embeddings as $x^{st}$ and $x^{wt}$. 
For each input embedding, the model  $f_{\theta}$  generates predicted probability  $z$  for each of the  $k$  classes.

In the subsequent sections, we detail the components of MUDAS.

\subsection{Pre-Processing for Acoustic Data}
\noindent\textbf{Mel Spectrograms} In sound classification, audio signals are often transformed into mel spectrograms, which are 2D time-frequency representations resembling images. This representation enables the use of advanced image-processing techniques for analyzing audio data. However, traditional Python-based signal processing libraries are unsuitable for mote-scale devices due to their resource constraints. To address this, ARM provides CMSIS (Cortex Microcontroller Software Interface Standard)\cite{CMSIS}, a suite of optimized libraries for embedded systems. Using the FFT functions in CMSIS-DSP, raw audio waveforms can be efficiently converted into mel spectrograms, enabling on-device audio preprocessing without requiring high computational resources.

\vspace*{1mm}
\noindent\textbf{Weak/Strong Augmentation} To enhance generalization, we apply weak augmentations such as time-reversing (x-axis flipping) and implement strong augmentations via SpecAugment \cite{SpecAugment}, which randomly masks portions of the time and frequency axes. These strategies compel the model to learn robust, invariant features, significantly improving its ability to generalize across diverse sound domains.

\vspace*{1mm}
\noindent\textbf{Feature Embeddings} After augmentation, we extract feature embeddings for urban sound classification, a method well-suited for the limited datasets in this domain. Unlike AdaMatch, which uses ResNet\cite{ResNet} for image-based tasks, our approach leverages embeddings specifically tailored for sound data. This not only improves generalization but also provides more interpretable representations of data relationships. By utilizing pre-trained models, we reduce the need for large datasets while maintaining efficiency through dimensionality reduction. For embedding generation, we use the SONYC-L\textsuperscript{3} model, optimized for efficient operation on resource-constrained devices.

\vspace*{-10pt}

\subsection{Training Classifier Model with Feature Embeddings}

\vspace*{1mm}
\noindent\textbf{Overview}
The process described for multi-label sound classification using MUDAS involves several steps. Feature embeddings from both source and target domains are processed together to update normalization statistics, which helps improve classifier performance. Unlike multi-class classification, which selects a single class, \emph{multi-label classification} uses sigmoid activation to output independent probabilities for each class, allowing multiple classes to be relevant simultaneously. \emph{Random probability interpolation} enhances feature diversity by combining probabilities from both batches using a random factor. \emph{Distribution alignment} involves creating a pseudo-label by merging source and target labels, with values capped at 1 to maintain class independence. \emph{Class-specific thresholding} assigns a threshold to each class based on its distribution in the source domain, improving detection sensitivity and providing flexibility in classification. Finally, \emph{diversity regularization} helps balance attention between present and absent labels, promoting robustness and preventing overfitting to any single class.


\vspace*{1mm}
\noindent\textbf{Batch Normalization and Downstream Classifier Model}
During the pre-processing, feature embedding batches are created from audio waveform data gathered from two distinct domains. The first batch includes embeddings  $x^{ss}$,  $x^{ws}$,  $x^{st}$, and  $x^{wt}$, while the second batch contains only  $x^{ss}$  and  $x^{ws}$. During probability generation, batch normalization statistics are updated exclusively with the first batch, which includes both source and target domain embeddings. 
After batch normalization, the two batches are, as shown in Figure \ref{fig:flow}, fed into the same downstream classifier model.
And we obtain the source probability values between the two batches, denoted as  $z^{ss'}$ and  $z^{ws'}$ from the first batch and  $z^{ss''}$ and $z^{ws''}$ from the second batch from the output of the classifer model.

In multi-label classification, unlike multi-class classification that uses a softmax activation to select a single class, each label is treated independently. A sigmoid activation is applied to each output node, allowing binary decisions across non-exclusive classes, with all class probabilities ranging between 0 and 1.

\vspace*{1mm}
\noindent\textbf{Random Probability Interpolation} To improve the diversity and robustness of feature representation, we employ random probability interpolation by combining probabilities from both batches. This method is inspired by the random logit interpolation used in AdaMatch, but instead of logits, we work with probabilities. This choice stems from the fact that, unlike softmax, the sigmoid activation function outputs independent class probabilities for each label. Each class probability is then interpolated using a unique random factor, $\lambda$, drawn from the range [0, 1]. This technique helps create a more varied and comprehensive representation across domains.

\vspace{-0.5em}
\begin{equation}
z^{ss}= \lambda \cdot z^{ss'} + (1-\lambda) \cdot z^{ss''}
\end{equation}
\vspace{-0.5em}
\begin{equation}
z^{ws}= \lambda \cdot z^{ws'} + (1-\lambda) \cdot z^{ws''}
\end{equation}

\noindent\textbf{Distribution Alignment} The aggregated pseudo-label $\hat{y}^{wt}$ is constructed by combining source and target pseudo-labels, enhancing the model’s adaptability to domain shifts. While this approach aligns with the core idea of AdaMatch, we deviate by avoiding normalization, which can compromise the independence of each class’s information. Instead, a minimum function is applied to cap any value exceeding 1, ensuring that class-specific distinctions are preserved while maintaining robust performance. 

\vspace{-0.5em}
\begin{equation}
\tilde{y}^{wt}_{(i,j)}= min(z^{wt}_{(i,j)} \cdot \frac{\frac{1}{n}\sum_{j=1}^{n}z^{ws}_{(i,j)}}{\frac{1}{m}\sum_{j=1}^{m}z^{wt}_{(i,j)}}, 1)
\end{equation}

\vspace*{1mm}
\noindent\textbf{Class-specific Thresholds}~~ While relative confidence thresholding, as used in AdaMatch, performs well in multi-class classification, it falls short for multi-label contexts. In multi-class settings, selecting the highest-probability class is effective due to the exclusivity of predictions. However, this method risks significant information loss in multi-label classification, where multiple classes may simultaneously hold relevant information.

To overcome this limitation, we introduce a class-specific thresholding approach, where each class is assigned an individualized threshold derived from its probability distribution in the source domain. This allows each class to be evaluated independently. Additionally, we employ dual thresholds—positive and negative—to improve detection sensitivity, capturing essential signals even for classes with lower probability values. This strategy preserves critical information, resulting in more balanced and precise multi-label predictions.

\vspace{-0.5em}
\begin{equation}
c^{+}_i= \tau^{+}\cdot\max_{j\in[1..n]} z^{ws}_{(i,j)}
\label{eq:pos_threshold}
\end{equation}
\vspace{-0.5em}
\begin{equation}
c^{-}_i= 1 - \tau^{-}\cdot(1 - \min_{j\in[1..n]} z^{ws}_{(i,j)})
\label{eq:neg_threshold}
\end{equation}

Let $c^{+}_i$ and $c^{-}_i$ represent the class-specific positive and negative thresholds for class $i$. Here, $\tau^{+}$ and $\tau^{-}$ are user-defined parameters for the positive and negative thresholds, respectively. If  $c^{+}_i$  is calculated to be smaller than  $c^{-}_i$  based on  $\tau^{+}$  and  $\tau^{-}$ , we swap the values to ensure that  $c^{+}_i$  represents the larger threshold and  $c^{-}_i$  the smaller one.

To compute these thresholds $c^{+}_i$ is the product of user-defined parameter for positive threshold $\tau^{+}$ and the maximum source probability $z^{ws}_{(i,j)}$ from the source domain for class $i$ over all batches $j$. $c^{-}_i$ is derived using user defined parameter for negative threshold $\tau^{-}$, where it scales the complement of the minimum source probability  $z^{ws}_{(i,j)}$ for class $i$ in all batches $j$.



Two source losses, $l^{ws}$ and $l^{ss}$, handle weak and strong augmentations, respectively. Complementing these, the two target losses—$l^{+}$ for positive classes and $l^{-}$ for negative classes—refine the model’s performance on the target domain by reinforcing correct classifications and reducing uncertainty in label assignment.

\vspace{-0.5em}
\begin{equation}
l^{ws} = -\frac{1}{k}\frac{1}{n}\sum_{i=1}^{k}\sum_{j=1}^{n} [y^{s}_{(i,j)} \cdot log z^{ws}_{(i,j)} - (1 - y^{s}_{(i,j)}) \cdot log (1 - z^{ws}_{(i,j)})] 
\end{equation}
\vspace{-0.5em}
\begin{equation}
l^{ss} = -\frac{1}{k}\frac{1}{n}\sum_{i=1}^{k}\sum_{j=1}^{n} [y^{s}_{(i,j)} \cdot log z^{ss}_{(i,j)} - (1 - y^{s}_{(i,j)}) \cdot log (1 - z^{ss}_{(i,j)}) ]
\end{equation}

\vspace{-0.5em}
\begin{equation}
l^{+} = -\frac{1}{k}\frac{1}{m}\sum_{i=1}^{k}\sum_{j=1}^{m} [\mathds{1}(\tilde{y}^{wt}_{(i,j)} \geq c^{+}_{i}) \cdot log z^{st}_{(i,j)} ]
\end{equation}
\vspace{-0.5em}
\begin{equation}
l^{-} = -\frac{1}{k}\frac{1}{m}\sum_{i=1}^{k}\sum_{j=1}^{m} [\mathds{1}(\tilde{y}^{wt}_{(i,j)} \leq c^{-}_{i}) \cdot (1 - log z^{st}_{(i,j)})] 
\end{equation}


\vspace*{1mm}
\noindent\textbf{Diversity Regularization} To further address the challenge of dataset imbalance, we apply the Diversity Regularization approach proposed in \cite{Diversity_Regularization}. This method introduces two distinct regularization strategies—one for positive labels and another for negative labels—ensuring the model gives balanced attention to both present and absent labels. By preventing overconfidence in individual labels, these strategies encourage diverse and well-calibrated predictions, which is especially critical in multi-label classification tasks where multiple correct labels can coexist.

\vspace{-0.5em}
\begin{equation}
l^{+}_{div}= \frac{1}{k}\frac{1}{m}\sum_{i=1}^{k}\sum_{j=1}^{m} [\mathds{1}(\tilde{y}^{wt}_{(i,j)} \geq c^{+}_{i}) \cdot z^{st}_{(i,j)} log z^{st}_{(i,j)}]
\end{equation}
\vspace{-0.5em}
\begin{equation}
l^{-}_{div}= \frac{1}{k}\frac{1}{m}\sum_{i=1}^{k}\sum_{j=1}^{m} [\mathds{1}(\tilde{y}^{wt}_{(i,j)} \leq c^{-}_{i}) \cdot (1 - z^{st}_{(i,j)}) \cdot log (1 - z^{st}_{(i,j)})]
\end{equation}

\vspace*{1mm}
\noindent\textbf{Total Loss} The total loss function integrates several components: the source weakly augmented loss, source strongly augmented loss, target positive and negative losses, and diversity regularization for both positive and negative cases. It is expressed as:
\vspace{-0.5em}
\begin{equation}
L = (l^{ws} + l^{ss}) + t(l^{+} + l^{-} + l^{+}_{div} + l^{-}_{div}),
\end{equation}

where $t$ is a time-dependent weight factor.

Distinct initial weights are assigned to the source loss, target loss, and diversity regularization terms. The source loss is given higher emphasis initially by assigning it a larger weight, while the target and diversity losses start with lower weights. Over time, these weights are dynamically adjusted to gradually balance all loss terms equally, ensuring a smooth transition and improving model performance across domains.

%% file: sections/relearning.tex
For deploying MUDAS in IoT devices, it is essential to address their stringent computational resource constraints.
In this section, we present the strategies that MUDAS uses to accommodate the on-device resource limitations.

\subsection{Optimization in Relearning Algorithm}\label{sec:opt_relearning}
To address the resource constraints of IoT devices, the MUDAS algorithm incorporates several optimizations.

\noindent\textbf{Initialization with Pre-trained Networks}
Instead of training a model from scratch using augmented source domain data, we initialize our model with a pre-trained network from the source domain. This approach reduces both computational and storage demands while leveraging domain-specific knowledge to enhance performance.

\noindent\textbf{Efficient Feature Representation Reuse}
We enhance MUDAS's compute efficiency by reusing feature representations for the urban sounds context that are known to be robust and to generalize well across domains \cite{bello2019sonyc}. For embedding generation, we utilize the SONYC-L\textsuperscript{3} model, optimized for resource-constrained devices. This model reduces the melspectrogram’s granularity and decreases the number of convolutional filters, retaining performance close to its predecessor L\textsuperscript{3} model (whose static and dynamic memory needs exceed that of mote-scale IoT devices) and outperforming VGGish. Moreover, instead of retraining the entire model, we decompose it, focusing on generating stable feature embeddings, and only fine-tuning the downstream classifier for the target domain for adapting the model (cf.~Figure \ref{fig:structure}). 

\noindent\textbf{Intermittent Model Updates}
Continuous on-device training is impractical and unnecessary for domain adaptation. Instead, the model is updated intermittently—when sufficient target data has been collected or when inference confidence significantly changes. During these intervals, raw data is processed for real-time inference, and only the smaller feature embeddings are saved on the device, minimizing storage requirements. Moreover, as embeddings are generated during normal inference, no additional computation is required. Additionally, as discussed in the next subsection, not all embeddings need to be retained.


\begin{figure}[!t]
    \centering    
    \begin{minipage}[t]{.5\linewidth}
    \centering
    \includegraphics[width=0.9\linewidth, angle=-90]{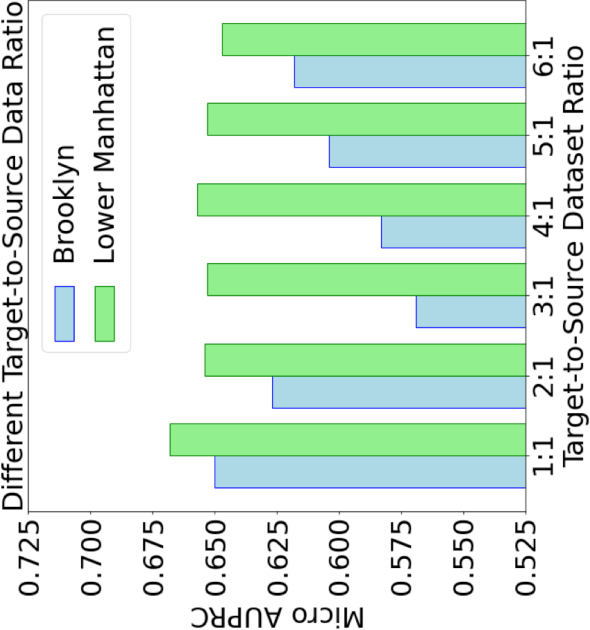}
    \label{fig:obj_diff_ratio}
    \end{minipage}%
    \begin{minipage}[t]{.5\linewidth}
    \centering
    \includegraphics[width=0.9\linewidth, angle=-90]{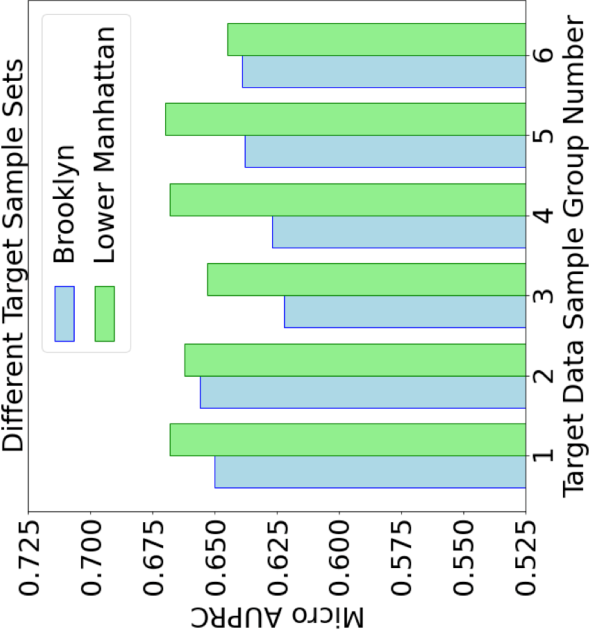}
    \label{fig:SC_passive}    \end{minipage}  
    \vspace{-4pt}
    \caption{Micro AUPRC performance of varying unlabeled target dataset sizes vs. consistent dataset size with different splits (3-class evaluation)}
    \vspace{-15pt}
    \label{fig:observation1}
\end{figure}

\setlength{\tabcolsep}{1pt}
\begin{table}[!b]
    \vspace{-10pt}
  \caption{The Micro AUPRC performance of MUDAS for target data samples categorized by probability ranges: high, mid, and low (3-class evaluation)}
  \vspace{-4pt}
  \label{tab:probability_target}
  \resizebox{8.5cm}{!}{%
  \begin{tabular}{ccccccc}
   \hline
      & &  \begin{tabular}[c]{@{}c@{}}Set with High\\ Probability\end{tabular}   & & \begin{tabular}[c]{@{}c@{}}Set with Medium\\ Probability\end{tabular}  & & \begin{tabular}[c]{@{}c@{}}Set with Low\\ Probability\end{tabular} \\
    \hline
      \begin{tabular}[c]{@{}c@{}}Brooklyn\end{tabular} & &  0.644 & &  0.601 & &   0.590  \\
    \hline
      \begin{tabular}[c]{@{}c@{}}Lower Manhattan\end{tabular}   & &  0.671 & &  0.646 & &  0.655  \\ 
  \hline
\end{tabular}}
\end{table}

\subsection{Selection for Data Used in Relearning}
MUDAS achieves memory and resource efficiency by simplifying its structure and fine-tuning only the downstream classifier. A key consideration is identifying which data is both useful and sufficient for this fine-tuning process.

Currently, MUDAS stores both source and target feature embeddings for processing. The storage of thousands of embeddings consumes several megabytes of flash memory, which represents a significant portion of the available memory on high-capability mote-scale platforms. To address these challenges, we are investigating strategies for selecting and optimizing the data storage requirements, based on our testing with MUDAS.

In our effort to refine the data selection process, we begin by sharing the key insights and observations we gained from these tests.

\vspace*{1mm}
\noindent\textbf{Key Observations from MUDAS Testing} Our simulation studies using the SONYC-UST dataset highlighted three key insights: we selected three classes—engine, alert signal, and human voice—that had relatively larger label distributions compared to the other classes. This selection was made to reduce the impact of dataset imbalance, allowing for clearer observations and more accurate analysis.

\begin{itemize}
    \item \textbf{Adequate adaptation gains from modest target dataset size} To assess the impact of target dataset size on accuracy, we conducted experiments with a fixed source dataset of 500 and varied target datasets: 500 (1:1), 1000 (2:1), 1500 (3:1), 2000 (4:1), 2500 (5:1), and 3000 (6:1), using two target locations and one source location. As shown in Figure \ref{fig:observation1} (left), increasing the target dataset size led to modest accuracy changes: Micro AUPRC fluctuated by 0.07 in Brooklyn and 0.02 in Lower Manhattan. 
    These results indicate that a \emph{modest target-to-source ratio, such as 1:1 or 2:1, is sufficient for effective adaptation and minimize performance degradation.}
    
    
    \vspace*{1mm}
    \item \textbf {Impact of label proportion in target datasets} To gain deeper insights, we conducted additional experiments with 500 source samples and 500 target samples, varying the target data splits while keeping the source data constant. As shown on the right side of Figure \ref{fig:observation1}, even with a constant number of target samples, Micro AUPRC accuracy fluctuated by approximately 0.03 at both locations. This variability led to further investigation, which revealed that the distribution of labels within the target dataset had a significant impact on performance. 
    
    \vspace*{1mm}
    \item \textbf {Impact of probability range} In real-world scenarios, label information may not always be available for the target dataset. Therefore, we explored the use of probability scores as an alternative source of information, under the assumption that higher probabilities are more likely to correspond to correctly labeled data. Our MUDAS framework leverages probability scores from classifier inference. We trained the model using only the source dataset and used it to infer probability scores for the target dataset. These probabilities were then normalized by dividing each class’s score by the range between its maximum and minimum values, creating three sets based on average probability: high, medium, and low. The 500 target samples were categorized accordingly. We tested MUDAS on these three target sets, and as shown in Table \ref{tab:probability_target}, the high-probability set consistently achieved significantly higher Micro AUPRC scores compared to the medium and low-probability sets. These results demonstrate that \emph{data selection based on high-probability target samples leads to better performance than selecting from medium or low-probability samples}.
    
\end{itemize}

These findings underscore the importance of data quality and label confidence in unsupervised adaptation tasks.

\vspace*{1mm}
\noindent\textbf{Data Selection Method of MUDAS} 
Based on three key observations, we present a heuristic method to selectively retain only high-confidence datasets.

\vspace{-0.5em}
\begin{equation}
D = \frac{1}{k}\sum_{i=1}^{k}(\frac{w_i \cdot z^{wt}_{i,j}}{\max_{j=1}^{m}z^{ws}_{i} - \min_{j=1}^{m}z^{ws}_{i}})
\end{equation}
\vspace{-0.3em}
The term  $z^{wt}_{i,j}$  represents the probability of the current data point  $j$  for class  $i$. After generating the target probability  $z^{wt}_{i,j}$, it is normalized using the range of the source dataset probabilities for each class, defined by  $\max_{j=1}^{m} z^{ws}_{i}$  and  $\min_{j=1}^{m} z^{ws}_{i}$. This normalization leverages the source dataset, which is reliably available in practice, especially since the target dataset may be incomplete or insufficient to accurately determine maximum and minimum values. Furthermore, since each class has its own probability range, normalization ensures consistency across the classes.

Once the normalization is applied, the target probability  $z^{wt}_{i,j}$  is multiplied by the class weight  $w_i = \frac{\text{Number of labels in class $i$}}{\text{Total number of all class labels}}$. This weighting assigns higher importance to classes with more frequent labels in the source dataset. The weighted values for each class are then combined and averaged to obtain the value $D$. Finally, data selection is completed by ranking the values in $D$  from highest to lowest.

When feature embeddings are generated at the target location, the decision to store them depends on the value of $D$. Initially, when there are insufficient feature embeddings, they are stored. However, as more embeddings are collected, only those with higher values than the lowest one are retained. The lowest-value embedding is removed to make room for the new one, ensuring that only the most relevant embeddings are stored.

%% file: sections/evaluation.tex
In this section, we evaluate the performance of MUDAS.
We use Version 2 of the SONYC-UST dataset \cite{SONYC_UST_V2}.  
This dataset provides latitude and longitude coordinates for each recording, allowing us to organize samples by specific locations across New York City. It includes 8 coarse-grained and 23 fine-grained sound classes; for this study, we focus on the broader coarse-grained classes to simplify the analysis and improve generalizability.

To simulate real-world domain shifts, we partition the dataset based on geographic location, designating one site as the source domain and a different site as the target domain. Both the target and validation datasets are sourced from the same location to maintain consistency in evaluation conditions. We selected Washington Square as the target domain due to its extensive data subset, comprising 306 out of a total of 538 validation recordings. To evaluate MUDAS’s adaptability across diverse environments, we systematically varied the source domain using data from four distinct locations: Union Square, Lower Manhattan, Brooklyn, and Central Park. 
In our evaluations, certain labels were excluded due to a lack of true samples. Specifically, the overarching classes ‘music’ and ‘dog’ were removed from all simulations. 
Our experimental setup incorporates 500 labeled samples from the source domain and 500 unlabeled samples from the target domain—a configuration that yielded suitable performance as detailed in Section 5. This approach allows for a thorough examination of MUDAS’s resilience and effectiveness in domain shift scenarios.

To enhance feature diversity, we apply both strong and weak data augmentations. Strong augmentation is achieved using SpecAugment \cite{SpecAugment}, which generates meaningful variations in the audio spectrograms, while weak augmentation consists of simple x-axis flip-flop transformations. 
Feature embeddings are generated using SONYC-L\textsuperscript{3} \cite{MKII}, a lightweight embedding model optimized for deployment on mote-scale devices, which ensures efficient processing for resource-constrained environments.

Our evaluation framework includes two baselines to benchmark MUDAS’s performance in the presence of domain shifts: (a) Lower Bound, where the model is trained on labeled source domain data and tested on target domain data, and (b) Upper Bound, where the model is trained and tested exclusively on labeled target domain data. These baselines allow us to quantify MUDAS’s effectiveness in adapting to new environments.

We evaluate MUDAS’s robustness across several scenarios: varying target locations, hyperparameter tuning (exploring different values for the positive threshold $\tau^+$ and negative threshold $\tau^-$, batch size), and conducting ablation studies to investigate the contributions of individual components to overall performance. Furthermore, we compare MUDAS with AdaMatch \cite{AdaMatch}, a framework designed for multi-class classification. Although this comparison is not an apples-to-apples comparison as the models are trained differently depending on the multi-label or the multi-class objective, it establishes two distinct baseline configurations: one for AdaMatch in multi-class classification settings and another for MUDAS in multi-label classification settings. This enables us to evaluate the relative improvements achieved by MUDAS, despite the differences in problem settings.

The feature embedding consists of 256 dimensions, and an MLP classifier with two hidden layers and dropout regularization is employed. The Adam optimizer is used, with the learning rate decreasing continuously based on cosine decay, starting from 0.001 and reducing to 0.00025. The batch size is set to 64, and the model is trained for 50 epochs.


\setlength{\tabcolsep}{1.5pt}
\begin{table}[!t]
  \caption{Performance of MUDAS in various source locations (6-class evaluation)}
    \vspace{-4pt}
  \label{tab:location}
  \resizebox{8.5cm}{!} 
  {\begin{tabular}{cccccc}
    \hline
    \multirow{2}{*}{Method} & \multirow{2}{*}{Metric} &  \multicolumn{4}{c}{Source Location} \\
    \cline{3-6}
    &   & \begin{tabular}[c]{@{}c@{}}Union\\ Square\end{tabular} & \begin{tabular}[c]{@{}c@{}}Central\\Park\end{tabular} & Brooklyn & \begin{tabular}[c]{@{}c@{}}Lower\\ Manhattan\end{tabular}      \\
    \hline
    \multirow{3}{*}{\begin{tabular}[c]{@{}c@{}}Lower-Bound\\ for MUDAS\end{tabular}} & Micro-AUPRC & 0.657 & 0.733 & 0.479 & 0.551     \\
    & F1-Score (0.5)   & 0.041  & 0.671 & 0.087 & 0.566     \\
    & Macro-AUPRC & 0.503  &  0.583 & 0.379 & 0.511     \\
    \hline
    \multirow{3}{*}{MUDAS} & Micro-AUPRC & 0.709 & 0.744 & 0.506 & 0.578  \\
    & F1-Score (0.5) & 0.041 & 0.690 & 0.149 & 0.609    \\
    & Macro-AUPRC & 0.528 & 0.570 & 0.390 & 0.532    \\
    \hline
    \multirow{3}{*}{\begin{tabular}[c]{@{}c@{}}Upper-Bound\\ for MUDAS\end{tabular}} & Micro-AUPRC & \multicolumn{4}{c}{0.778} \\
    & F1-Score (0.5) &  \multicolumn{4}{c}{0.706}   \\
    & Macro-AUPRC & \multicolumn{4}{c}{0.601} \\
  \hline
\end{tabular}}
\vspace{-10pt}
\end{table}

\begin{figure}[!b]
    \vspace{-5pt}
    \centering    
    \begin{minipage}[t]{.5\linewidth}
    \centering
    \includegraphics[width=0.67\linewidth, angle=-90]{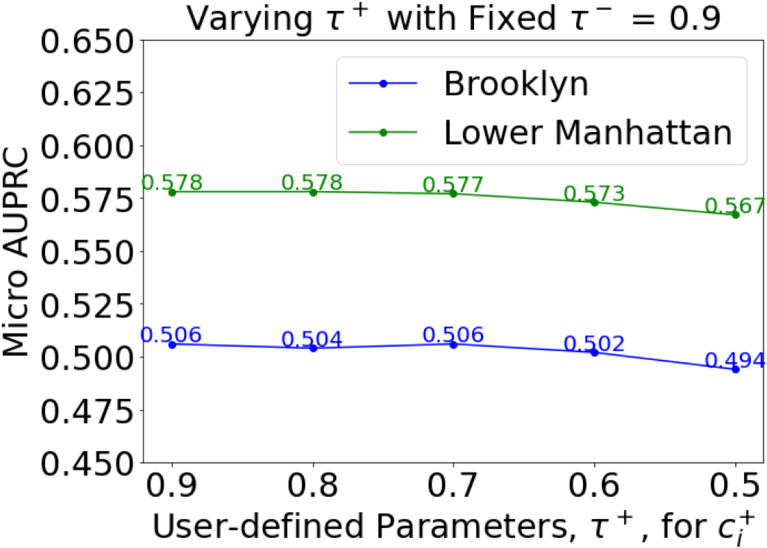}
    \label{fig:eval_pos_tau}
    \end{minipage}%
    \begin{minipage}[t]{.5\linewidth}
    \centering
    \includegraphics[width=0.67\linewidth, angle=-90]{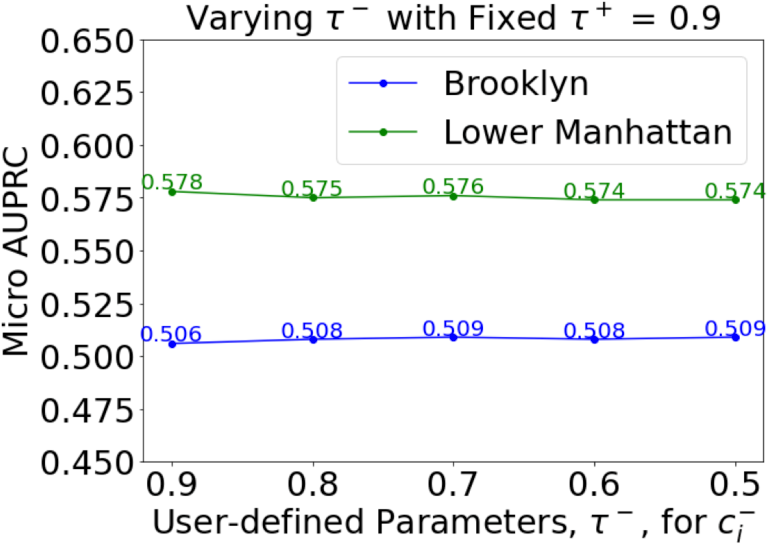}
    \label{fig:eval_neg_tau}    \end{minipage}  
    \vspace{-4pt}
    \caption{Impact of varying user-defined parameters for positive and negative thresholds, $\tau^+$ and $\tau^-$ (6-class evaluation)}
    \label{fig:eval_tau}
\end{figure}

\vspace*{1mm}
\noindent\textbf{Location-based Evaluation.} To evaluate MUDAS’s performance, we selected four distinct locations as source domains, each contributing 500 labeled samples for training, with Washington Square as the target domain providing 500 unlabeled samples. We set $\tau^+ = 0.9$ and $\tau^- = 0.9$ with a batch size of 64. MUDAS’s performance was compared against lower-bound and upper-bound baselines, as shown in Table~\ref{tab:location}.

The results highlight MUDAS’s adaptability in sound event classification. In terms of Micro-AUPRC, MUDAS outperforms the lower-bound baseline across all locations, with improvements ranging from 0.052 at Union Square to 0.011 at Central Park. These findings demonstrate the effectiveness of MUDAS in discriminating sound events in diverse and complex acoustic environments.

F1-Score results indicate significant gains in Brooklyn (0.62), Lower Manhattan (0.43), and Central Park (0.019), albeit Union Square shows no improvement. This suggests precision-recall trade-offs exist in certain locations.

For Macro-AUPRC, MUDAS consistently surpasses the lower-bound baseline at most locations, with increases of 0.025 at Union Square, 0.021 at Lower Manhattan, and 0.011 at Brooklyn, indicating its ability to manage class imbalances. However, a slight decrease of -0.013 at Central Park points to areas for further refinement, especially in handling location-specific factors.

Although MUDAS shows notable gains over the upper-bound baseline, a performance gap remains, especially in Micro-AUPRC and F1-Score. These results underscore opportunities for further optimization, particularly in achieving upper-bound-level performance in challenging settings.


\begin{table}[!t]
  \caption{Ablation Study: impact of negative threshold and diversity regularization in MUDAS Performance (Micro AUPRC) (6-classes evaluation)}
  \vspace{-4pt}
  \label{tab:setting}
  \resizebox{8.5cm}{!}{%
  \begin{tabular}{ccccccc}
   \hline
    \multirow{2}{*}{\begin{tabular}[c]{@{}c@{}}MUDAS w/\\ Positive\\ Threshold\end{tabular}}&\multirow{2}{*}{\begin{tabular}[c]{@{}c@{}}Negative\\ Threshold\end{tabular}} &\multirow{2}{*}{\begin{tabular}[c]{@{}c@{}}Diversity \\ Regularization\end{tabular}} & \multicolumn{4}{c}{ Source Location}\\
    \cline{4-7}
    & & & \begin{tabular}[c]{@{}c@{}}Union\\ Square\end{tabular} &  \begin{tabular}[c]{@{}c@{}}Central\\ Park\end{tabular} &  \begin{tabular}[c]{@{}c@{}}Brooklyn\\ \end{tabular} & \begin{tabular}[c]{@{}c@{}}Lower\\ Manhattan\end{tabular}\\
    \hline
    O & X & X & 0.686$\pm$.020 & 0.735$\pm$.005 & 0.482$\pm$.012 &  0.582$\pm$.003\\
    \hline
    O & O & X & 0.691$\pm$.019 & 0.736$\pm$.006 & 0.492$\pm$.015& 0.582$\pm$.004 \\
    \hline
    O & X & O & 0.693$\pm$.019 & 0.739$\pm$.005 & 0.493$\pm$.012 & 0.572$\pm$.004 \\
    \hline
    O & O & O & 0.695$\pm$.018 & 0.740$\pm$.007 & 0.499$\pm$.013 & 0.570$\pm$.006 \\
    \hline
\end{tabular}}
\vspace{-20pt}
\end{table}

\begin{figure}[!b]
    \vspace{-2pt}
    \centering    
    \begin{minipage}[t]{.5\linewidth}
    \centering
    \includegraphics[width=0.67\linewidth, angle=-90]{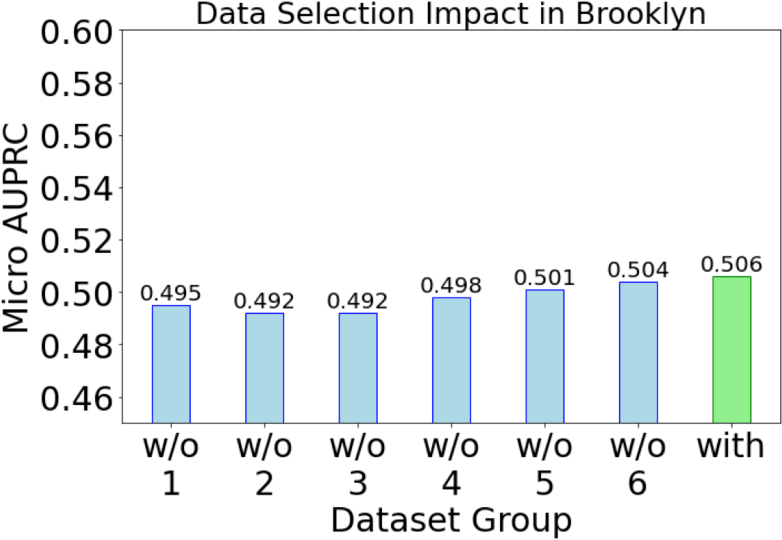}
    \label{fig:eval_ds_br}
    \end{minipage}%
    \begin{minipage}[t]{.5\linewidth}
    \centering
    \includegraphics[width=0.67\linewidth, angle=-90]{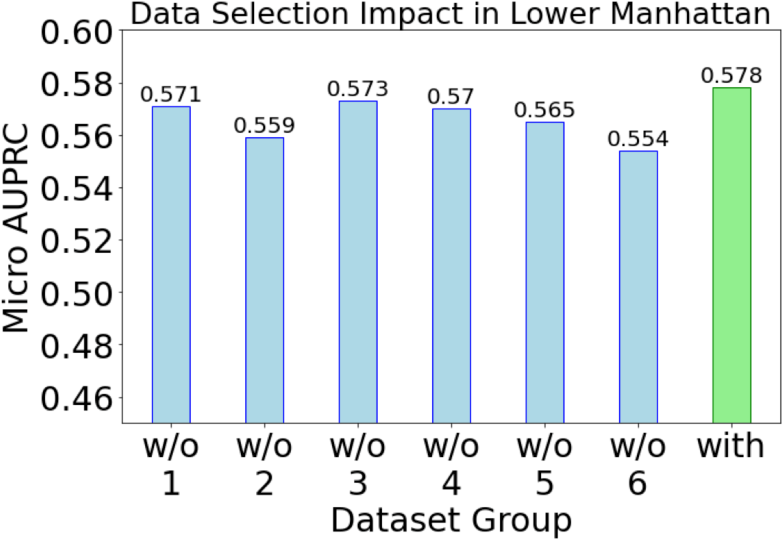}
    \label{fig:eval_ds_lm}    \end{minipage} 
    \vspace{-2pt}
    \caption{Ablation study: Impact of data selection on MUDAS performance (6-class evaluation)}
    \label{fig:eval_ds}
\end{figure}

\vspace*{1mm}
\noindent\textbf{Hyper-parameter Study.}
We begin by evaluating the user-defined parameters for the positive and negative thresholds, $\tau^+$ and $\tau^-$, to understand their impact on model performance. Initially, we fix $\tau^-$ at 0.9 and incrementally vary $\tau^+$ from 0.5 to 0.9 in steps of 0.1. Then, we fix $\tau^+$ at 0.9 and adjust $\tau^-$ from 0.5 to 0.9, also in 0.1 increments. It is crucial to note that lowering $\tau^+$ too much can result in overlap, where $\tau^+$ becomes smaller than $\tau^-$, requiring careful calibration to maintain a balance between high-confidence positive and negative pseudo-labels.

The results, shown in Figure~\ref{fig:eval_tau}, demonstrate the distinct effects of varying these thresholds. Lowering $\tau^+$ increases the amount of pseudo-labeled data by broadening the predicted labeled dataset, but this comes at the cost of reduced performance. For instance, decreasing $\tau^+$ from 0.9 to 0.5 leads to a Micro-AUPRC decrease of 0.12 for Brooklyn and 0.11 for Lower Manhattan. In contrast, reducing $\tau^-$ from 0.9 to 0.5 has a more subtle effect: while it initially causes a slight decline in Micro-AUPRC, it results in marginal improvements at the lowest threshold.

Overall, the analysis shows that performance is more sensitive to reductions in $\tau^+$ than $\tau^-$. This highlights the importance of carefully calibrating $\tau^+$ to avoid introducing excessive noise while ensuring an adequate amount of labeled data points.

\vspace*{1mm}
\noindent\textbf{Ablation Study.}
In the ablation study, we systematically analyzed the effects of incorporating negative thresholds, diversity regularization, and data selection by comparing model performance across different configurations. Table~\ref{tab:setting} presents the results for MUDAS, beginning with the baseline configuration that uses only a positive threshold and incrementally adding negative thresholds and diversity regularization.

The results demonstrate that the impact of negative thresholds and diversity regularization varies across locations. In Brooklyn, adding a negative threshold improves performance more significantly than using diversity regularization alone. In contrast, Central Park sees a greater improvement from diversity regularization compared to the negative threshold. However, in Lower Manhattan, diversity regularization slightly reduces accuracy, indicating that it may not always be beneficial, especially in environments with balanced class distributions or where the model already performs near optimally.

Overall, the combination of positive and negative thresholds with diversity regularization achieves the highest performance in most locations. These findings underscore the importance of integrating all three components to enhance the robustness and adaptability of the MUDAS framework.

Regarding data selection, shown in Figure~\ref{fig:eval_ds}, the results were largely consistent with those observed without data selection, yet we noted significant improvements in Micro-AUPRC scores for both Brooklyn and Lower Manhattan. These improvements suggest that data selection plays a key role in identifying beneficial unlabeled datasets, which can refine model performance.

\setlength{\tabcolsep}{1pt}
\begin{table}[!t]
  \centering
  \caption{Performance comparison between AdaMatch and the baseline in multi-class classification settings (3-class evaluation)}
  \vspace{-4pt}
  \label{tab:method_AdaMatch}
  \large
  \resizebox{8.5cm}{!}{%
  \begin{tabular}{lccccc}
    \hline
     & & \multicolumn{4}{c}{Source Location} \\
    \cline{3-6} 
     & & \begin{tabular}[c]{@{}c@{}}Union\\ Square\end{tabular} &  \begin{tabular}[c]{@{}c@{}}Central\\ Park\end{tabular} &  \begin{tabular}[c]{@{}c@{}}Brooklyn\\ \end{tabular} & \begin{tabular}[c]{@{}c@{}}Lower\\ Manhattan\end{tabular}\\
    \hline 
    Metric & \multicolumn{5}{c}{} \\
    \hline
    \multirow{2}{*}{Micro-AUPRC} & Baseline &  0.732\small$\pm$.007 & 0.738\small$\pm$.007 & 0.646\small$\pm$.006 & 0.661\small$\pm$.009 \\
	& AdaMatch & 0.749\small$\pm$.008 & 0.741\small$\pm$.004 & 0.666\small$\pm$.008 & 0.675\small$\pm$.008 \\
    \hline 
    Class Label & \multicolumn{5}{c}{} \\
    \hline
    \multirow{2}{*}{Engine} & Baseline &  0.710\small$\pm$.006 & 0.716\small$\pm$.006 &  0.651\small$\pm$.004 & 0.662\small$\pm$.004\\
	& AdaMatch & 0.725\small$\pm$.006 & 0.718\small$\pm$.004 &  0.661\small$\pm$.007 & 0.667\small$\pm$.006 \\
    \hline
    \multirow{2}{*}{Alert signal} & Baseline & 0.776\small$\pm$.016 & 0.758\small$\pm$.012 &  0.339\small$\pm$.013 & 0.705\small$\pm$.013 \\
	& AdaMatch & 0.774\small$\pm$.010 & 0.760\small$\pm$.011 & 0.353\small$\pm$.020 & 0.703\small$\pm$.006\\
    \hline
    \multirow{2}{*}{Human voice} & Baseline &0.838\small$\pm$.003 & 0.877\small$\pm$.006 & 0.826\small$\pm$.015 & 0.853\small$\pm$.017 \\
    & AdaMatch & 0.847\small$\pm$.006 & 0.876\small$\pm$.006 &  0.853\small$\pm$.010 & 0.855\small$\pm$.013  \\
    \hline
  \end{tabular}%
  }  
  \vspace{-10pt}
\end{table}

\vspace*{1mm}
\noindent\textbf{Evaluation of AdaMatch and MUDAS Performance Across Multi-Class and Multi-Label Settings.}
Although MUDAS is built upon the AdaMatch framework, a direct comparison is not straightforward due to fundamental differences in task formulation. AdaMatch is designed for multi-class classification, utilizing softmax activation and categorical cross-entropy loss, while MUDAS focuses on multi-label classification, employing sigmoid activation and binary cross-entropy loss. 

To assess MUDAS's and AdaMatch’s adaptability in a multi-label context, we establish two separate baseline configurations: one for multi-class classification (AdaMatch) and one for multi-label classification (MUDAS). To evaluate the performance of each method in isolation, two different baseline approaches are employed. The first baseline operates in a multi-class classification setting, utilizing a softmax activation function and categorical cross-entropy as the loss function. The second baseline is designed for a multi-label classification setting, employing a sigmoid activation function and binary cross-entropy for the loss function. Both baselines are trained exclusively on source-labeled data, providing a benchmark for evaluating the proposed methods. 

In this evaluation, we set $\tau = 0.9$ for AdaMatch and $\tau^+ = 0.9$, $\tau^- = 0.9$ for MUDAS, maintaining a consistent batch size of 64 for both models.
To reduce measurement bias due to class imbalance---wherein multi-class classification often does not predict underrepresented classes, leading to artificially inflated performance---we limit the set of classes to those that have a sufficient number of labels. Accordingly, 3 classes are selected for a fairer comparative evaluation, namely engine, alert signal, and human voice. 

We evaluated the performance of the models using data from four different locations, with the results presented in Table \ref{tab:method_AdaMatch}, \ref{tab:method_MUDAS}. The evaluation shows that AdaMatch consistently outperforms its baseline for multi-class classification, while MUDAS consistently outperforms its baseline for multi-label classification. Relatively speaking, AdaMatch shows improvements of up to 0.020, whereas MUDAS shows stronger improvements of up to 0.067. These results corroborate that both AdaMatch and MUDAS are effective unsupervised domain transfer techniques.



The inversion observed in Brooklyn underscores the importance of appropriately selecting and sizing the dataset. 
While urban sound classification is fundamentally a multi-label problem, the superior performance of the multi-class baseline and AdaMatch in Brooklyn suggests that the dataset or training process may not have been adequate for this location.
Indeed, Brooklyn demonstrates the lowest accuracy among the four locations, pointing to potential issues in the model training process. 
This is corroborated by experiments we did with improved data quality, where MUDAS and its multi-label baseline show improvements at a faster rate than they do for AdaMatch. We thus expect MUDAS and the multi-label baseline to outperform their multi-class counterparts, aligning with their superior performance observed in other locations.

\setlength{\tabcolsep}{1pt}
\begin{table}[!t]
  \centering
  \caption{Performance comparison between MUDAS and the baseline in multi-label classification settings (3-class evaluation)}
  \vspace{-4pt}
  \label{tab:method_MUDAS}
  \large
  \resizebox{8.5cm}{!}{%
  \begin{tabular}{lccccc}
    \hline
     & & \multicolumn{4}{c}{Source Location} \\
    \cline{3-6} 
     & & \begin{tabular}[c]{@{}c@{}}Union\\ Square\end{tabular} &  \begin{tabular}[c]{@{}c@{}}Central\\ Park\end{tabular} &  \begin{tabular}[c]{@{}c@{}}Brooklyn\\ \end{tabular} & \begin{tabular}[c]{@{}c@{}}Lower\\ Manhattan\end{tabular}\\
    \hline 
    Metric & \multicolumn{5}{c}{} \\
    \hline
    \multirow{2}{*}{Micro-AUPRC} & Baseline &  0.763\small$\pm$.014 & 0.824\small$\pm$.007 &  0.575\small$\pm$.020 & 0.665\small$\pm$.006 \\
	& MUDAS & 0.801\small$\pm$.005 & 0.826\small$\pm$.005 & 0.642\small$\pm$.010 & 0.668\small$\pm$.007  \\
    \hline 
    Class Label & \multicolumn{5}{c}{} \\
    \hline
    \multirow{2}{*}{Engine} & Baseline &  0.744\small$\pm$.018 & 0.856\small$\pm$.008 & 0.627\small$\pm$.012 &  0.693\small$\pm$.013\\
	& MUDAS & 0.790\small$\pm$.020 & 0.851\small$\pm$.010 & 0.681\small$\pm$.007 & 0.682\small$\pm$.005 \\
    \hline
    \multirow{2}{*}{Alert signal} & Baseline &  0.749\small$\pm$.012 & 0.756\small$\pm$.009 & 0.280\small$\pm$.005 &  0.651\small$\pm$.041\\
	& MUDAS & 0.749\small$\pm$.019 & 0.772\small$\pm$.012   & 0.267\small$\pm$.019 & 0.714\small$\pm$.010\\
    \hline
    \multirow{2}{*}{Human voice} & Baseline & 0.856\small$\pm$.012 & 0.873\small$\pm$.005 & 0.679\small$\pm$.030 & 0.839\small$\pm$.021 \\
    & MUDAS & 0.875\small$\pm$.017 & 0.864\small$\pm$.005  &  0.768\small$\pm$.002 & 0.849\small$\pm$.017 \\
    \hline
  \end{tabular}
  }  
  \vspace{-10pt}
\end{table}

\vspace*{1mm}
\noindent\textbf{Feasibility of Deployment on Edge Devices}
Although MUDAS has been evaluated in an emulated environment, the derived model can be readily deployed on existing edge devices. Each 1-second embedding, which would be sufficient for accurate noise classification, needs only 1 KB. Thus, when using 500 source and target embeddings, the total memory requirement is just over 2 MB of flash storage. Moreover, the model utilizes only 141 KB of activation memory and 67 KB of parameter memory during training, highlighting its suitability for deployment on resource-constrained edge platforms.


%% file: sections/conclusion.tex
In this work, we introduce MUDAS, a novel framework tailored for multi-label classification in resource-constrained IoT environments under UDA. MUDAS addresses the unique challenges of multi-label scenarios and overcomes the limitations of traditional UDA methods by integrating class-specific adaptive thresholds and diversity regularization. These techniques significantly enhance pseudo-labeling accuracy and model robustness. Additionally, MUDAS employs a selective retraining strategy that ensures computational efficiency, making it highly suitable for deployment on mote-scale IoT devices with stringent resource constraints.

Extensive evaluations using the SONYC-UST dataset validate MUDAS’s ability to handle domain shifts effectively, achieving superior performance compared to baseline models for multi-label urban sound classification. Importantly, MUDAS achieves this while adhering to the practical limitations of real-world IoT environments. These results underscore MUDAS’s potential to scale and adapt UDA methods for complex and dynamic settings.
Note that our evaluation of MUDAS assumes training from scratch. However, in practical scenarios, as described in Section~\ref{sec:opt_relearning}, MUDAS ought to be initialized with a pre-trained source model, which would further reduce the reliance on extensive source datasets and save valuable storage space on resource-constrained devices.

While our study demonstrates the effectiveness of MUDAS in sound classification, the framework has the potential to generalize to other multi-label domains. Future research will explore its application to diverse classification problems to broaden its utility.

This work serves as foundation for future research in multi-label domain adaptation and resource-efficient learning on edge devices, emphasizing the critical balance between adaptation performance and the practical constraints of deployment.  Although we have intentionally limited the learning to the edge device itself in this paper, there are contexts where coordination of the learning with neighboring devices is warranted. We thus envision extending MUDAS to federated learning paradigms, where model updates are informed by unlabeled data from neighboring nodes. This decentralized approach has the potential to accelerate convergence to robust models by leveraging local collaboration, and enhance the system’s adaptability to dynamic environments.



